\documentclass{article}

\usepackage[square,numbers]{natbib} %[round]
\usepackage[preprint]{neurips_2024}

\usepackage[utf8]{inputenc} % allow utf-8 input
\usepackage[T1]{fontenc}    % use 8-bit T1 fonts
\usepackage{hyperref}       % hyperlinks
\usepackage{url}            % simple URL typesetting
\usepackage{booktabs}       % professional-quality tables
\usepackage{amsfonts}       % blackboard math symbols
\usepackage{nicefrac}       % compact symbols for 1/2, etc.
\usepackage{microtype}      % microtypography
\usepackage{xcolor}         % colors

\usepackage{amsmath,amsfonts}
\usepackage{amssymb}
\usepackage{graphicx}
\usepackage{url}

\usepackage{algorithmic}
\usepackage{graphicx}
\usepackage{graphics}
\usepackage{textcomp}
\usepackage{subfigure}
\usepackage{tabularx}
\usepackage{times}
\usepackage{multirow}
\usepackage{bm}
\usepackage{latexsym}
\usepackage{threeparttable}
\usepackage{epsf}
\usepackage{algorithm}
\usepackage{color,colortbl}

\title{UDQL: Bridging The Gap between MSE Loss and The Optimal Value Function in Offline Reinforcement Learning}

\author{%
 \textsuperscript{1, 2}Yu Zhang\thanks{Equal contribution.}, \  \textsuperscript{3}Rui Yu\footnotemark[1], \  \textsuperscript{1}Zhipeng Yao, \ \textsuperscript{2}Wenyuan Zhang, \ \textsuperscript{1}Jun Wang\thanks{Corresponding author.}, \ \textsuperscript{2}Liming Zhang\footnotemark[2] \ \\
  \textsuperscript{1}Shenyang University of Chemical Technology, \ \textsuperscript{2}University of Macau \ \textsuperscript{3}University of Louisville\\
 \texttt{zhangy@syuct.edu.cn}, \ \texttt{rui.yu@louisville.edu}, \ \texttt{yiucp@outlook.com}, \\ 
\texttt{yc07421@um.edu.mo}, \ \texttt{wangjun@syuct.edu.cn}, \ \texttt{lmzhang@um.edu.mo}\\ 
}
\begin{document}

\maketitle

\begin{abstract}
  The Mean Square Error (MSE) is commonly utilized to estimate the solution of the optimal value function in the vast majority of offline reinforcement learning (RL) models and has achieved outstanding performance. However, we find that its principle can lead to overestimation phenomenon for the value function. In this paper, we first theoretically analyze overestimation phenomenon led by MSE and provide the theoretical upper bound of the overestimated error. Furthermore, to address it, we propose a novel Bellman underestimated operator to counteract overestimation phenomenon and then prove its contraction characteristics. At last, we propose the offline RL algorithm based on underestimated operator and diffusion policy model. Extensive experimental results on D4RL tasks show that our method can outperform state-of-the-art offline RL algorithms, which demonstrates that our theoretical analysis and underestimation way are effective for offline RL tasks.
\end{abstract}

\section{Introduction}

Currently, many offline reinforcement models estimate the optimal value function ($V^*(s)$, $Q^*(s,a)$) and policy ($\pi^*(a|s)$) by finding $\max_{a} Q(s,a)$. In particular, IQL (implicit Q-learning) \citep{IQL} creatively proposes that it estimates $V(s)$ to represent the supremum of $Q(s,a)$ (the maximum $Q$-value under state $s$) by expectile regression $L^{\tau} _{2}$ as the Eq. (\ref{e1}). The expectile regression is closely associated with quantile regression \citep{Quantile}. IQL has achieved great success in the offline reinforcement learning by using $V(s)$ to estimate $\max_{a} Q(s,a)$ without out-of-distribution (OOD) actions. Many advanced offline reinforcement models following the track of IQL have been proposed to learn the optimal policy by estimating $\max_{a} Q(s,a)$ by $V(s)$ with policy constraints. However, we find that there are two problems in these methods.
\begin{equation}
	\begin{split}
		\label{e1}
		V(s) = \max_{a} Q(s,a).
	\end{split} 
\end{equation}
For one thing, the excellent work of \citep{XQL} reveals the fact that the errors existing in $max$ operator of reinforcement learning can be considered to be Gumbel distribution and then finding $\max_{a} Q(s,a)$ is modeled as sampling in Gumbel distribution based on Extreme Value Theorem (EVT) \citep{EVT1} \citep{EVT2}. For Gumbel distribution \citep{Gumbel} $\mathcal{G}(\mu, \beta)$, $\mu$ is its location parameter (mode) and $\beta$ is its scale parameter \citep{Gumbel1} \citep{Gumbel2}. Concretely, with respect to the estimated random variables $\hat{Q} (s,a)$ for real $Q(s,a)$ and $\hat{V} (s)$ for real $V(s)$, we get the Eq. (\ref{e21}) and (\ref{e22}) \citep{XQL} under Gumbel distribution.
\begin{align}
	\begin{split}
		\label{e21}
		&V(s_{t+1}) = \beta \log{\mathbb{E}_{a_{t+1} \sim \mu(\cdot |s_{t+1}) }\left ( \exp \left (Q(s_{t+1},a_{t+1})/\beta \right ) \right )  } := \mathbb{L}^{\beta}_{a_{t+1}}\left [ Q(s_{t+1},a_{t+1}) \right ], \\
		&Q(s_{t} ,a_{t}) = r(s_{t} ,a_{t}) + \gamma \max_{a_{t+1}}Q(s_{t+1},a_{t+1}) = r(s_{t} ,a_{t}) + \gamma \mathbb{L}^{\beta}_{a_{t+1}}\left [ Q(s_{t+1},a_{t+1}) \right ],
	\end{split} 
\end{align}
where $\mu(\cdot |s_{t+1})$ is the policy that generate the dataset.

\begin{align}
	\begin{split}
		\label{e22}
		&\hat{V} (s_{t+1}) \sim \mathcal{G}\left ( \mathbb{L}^{\beta}_{a_{t+1} }\left [ Q(s_{t+1},a_{t+1}) \right ] , \beta \right ), \\
		&\hat{Q} (s_{t} ,a_{t}) \sim \mathcal{G}\left ( r(s_{t} ,a_{t}) + \gamma \mathbb{L}^{\beta}_{a_{t+1}}\left [ Q(s_{t+1},a_{t+1}) \right ], \gamma \beta   \right ), 
	\end{split} 
\end{align}
where $t$ is the $t$-th timestep in the trajectory; $V(s)$ is optimal state value function and $Q(s,a)$ is optimal action value function; $\hat{V}(s)$ is the actual learned state value function and $\hat{Q}(s,a)$ is the actual learned action value function.

Furthermore, Mean Square Error (MSE) is always used to compute $\hat{V} (s)$ and $\hat{Q} (s,a)$ in most offline reinforcement models with optimal value function like IQL \citep{IQL}. However, there is the diverge between optimal value function and its computation means of MSE. It can be proved that the expectation of $\hat{Q} (s,a)$ is the optimal solution of $Q(s,a)$ for MSE loss $E (\hat{Q} (s,a)) = Q(s,a)$ and also the expectation of $\hat{V} (s)$ is the optimal solution of $V(s)$ for MSE loss $E (\hat{V} (s)) = V(s)$. Specifically, based on the Eq. (\ref{e1}) and the feature of Gumbel distribution, we can get the Eq. (\ref{e31}) and (\ref{e32}) because it is acknowledged that the expectation is the optimal solution for MSE.
%and policy such as
\begin{align}
	%\begin{split}
	%\label{e3}
	&\tilde{V} (s_{t+1}) = \mathbb{E} \left ( \hat{V} (s_{t+1}) \right ) = \mathbb{L}^{\beta}_{a_{t+1}}\left [ \tilde{Q}(s_{t+1},a_{t+1}) \right ] + \gamma_{e} \ast \beta \label{e31},\\
	&\tilde{Q} (s_{t} ,a_{t}) = \mathbb{E} \left ( \hat{Q} (s_{t} ,a_{t}) \right ) = r(s_{t} ,a_{t}) + \gamma \mathbb{L}^{\beta}_{a_{t+1}}\left [ \tilde{Q}(s_{t+1},a_{t+1}) \right ] + \gamma_{e} \ast  \gamma \beta \label{e32},
	%\end{split} 
\end{align}
where $\gamma$ is discount factor and $\gamma_{e}$ is Euler-Mascheroni constant. $\tilde{V}(s)$ and $\tilde{Q}(s,a)$ are optimal solutions of the state value function and the action value functions under MSE loss respectively.

Therefore, the divergence derives from the difference between expectation and location of Gumbel distribution while the expectation of Gaussian distribution has the same value with its mode. Concretely, the expectation of Gumbel distribution overestimates its location parameter and with regard to the previous state $s_t$ MSE gives the overestimated location parameters as the Eq.(\ref{e4}). So when MSE is exploited to estimate the solution of $\max_{a} Q(s,a)$, the overestimation of $Q$-value naturally occurs and we will give its theoretical bound in Section 3.

\textbf{Remark 1.0} The overestimated error $\gamma_{e} \ast  \gamma \beta$ is the nested error. In other words, $\hat{Q}(s_{t+1},a_{t+1})$ also has the overestimated error $\gamma_{e} \ast \beta$ and it backpropagates to the $Q$-value of the previous state-action $\hat{Q}(s_{t},a_{t})$.
\begin{equation}
	\begin{split}
		\label{e4}
		&\hat{V} (s_{t}) \sim \mathcal{G}\left ( \mathbb{L}^{\beta}_{a_{t} }\left [ \hat{Q}(s_{t},a_{t}) \right ] + \gamma_{e} \ast  \gamma \beta , \gamma\beta \right ), \\
		&\hat{Q} (s_{t} ,a_{t}) \sim \mathcal{G}\left ( r(s_{t} ,a_{t}) + \gamma \mathbb{L}^{\beta}_{a_{t+1}}\left [ \hat{Q}(s_{t+1},a_{t+1}) \right ] + \gamma_{e} \ast  \gamma \beta, \gamma \beta   \right ).
	\end{split}
\end{equation}
For another, if $\max_{a} Q(s,a)$ is used to represent $V$-value function, the corresponding policy must be the optimal one $\pi^{*}(s)$, which would be consistent in the value function and policy. However, in practice it is almost impossible to learn the optimal policy $\pi^{*}(s)$ in the offline dataset environment. Then the optimal value function $Q(s,a)$ and $V(s)$ have been used but the policy being used is not the corresponding optimal. So $\hat{V}(s)$ and $\hat{Q}(s,a)$ also have been overestimated because the policy cannot match the optimal action.
In addition, it is generally acknowledged that the operator $max$ will always overestimate the value functions.

In this work, we propose a novel Bellman backup operator to address overestimation problem of action state value function and then propose a new offline reinforcement learning algorithm based on that operator. Specifically, firstly we theoretically analyze the overestimation phenomenon of value functions under Gumbel distribution model in details. The quantity of estimated error is given under theory analysis. Furthermore, a novel Bellman operator is proposed to backup $Q$-value function based on expectile regression but it underestimates $Q(s,a)$ to counteract overestimation phenomenon instead of estimating $\max Q(s,a)$ like IQL series \citep{IQL} \citep{SRPO} \citep{IDQL} \citep{SfBC}. So it works in the under estimating way (not $max$ but $under$). Then we theoretically prove that it is a contraction operator. Additionally, we realize it by modifying the loss function to backup $Q(s,a)$ in SARSA-TD way. At last, we propose a offline reinforcement model based the proposed operator and diffusion policy, and we call it UDQL (Underestimated Diffusion-QL). Concretely, we estimate the value function by counteracting overestimation and employ the diffusion approach \citep{diffusionQL} \citep{CEP} \citep{EDP} \citep{SRPO} \citep{IDQL} \citep{SfBC} to learn the deterministic policy \citep{DPG} \citep{DDPG} \citep{TD3} \citep{D4PG} \citep{TD3BC}. More importantly, we find that when the value function is underestimated appropriately, the behavior cloning component is not necessary to the policy in some cases. 

We evaluate the performance of our algorithm on the popular offline reinforcement learning benchmark D4RL \citep{d4rl}. It can outperform the prior state-of-the-art offline reinforcement algorithms in gym-locomotion, Kitchen and Adroit tasks. In AntMaze tasks, especially the medium and large ones, we find that the value function need more powerful capability so as to learn the correct solution under underestimation condition and our method also can outperform the prior state-of-the-art algorithms.

%\citep{DDPG}  \citep{TD3} \citep{D4PG} to based on the work of \citep{diffusionQL}. without behavior cloning (BC) constraints

\section{Related Work}

Distributional shift is one of two primary reasons for performance degradation of RL models in offline settings. The other challenge is extrapolation errors that stem from overestimation of out-of-distribution (OOD) actions and that are exacerbated by bootstrapping \citep{BEAR}. In order to address them, many offline RL algorithms employ policy regularization or constraints to restrict the policy from taking out-of-distribution (OOD) action. For example, behavior cloning (BC) \citep{BC} is the main constraint approaches to make the trained policy imitate the behavior that generates the offline dataset \cite{BCQ} \citep{BRAC} \citep{TD3BC} \citep{LatentAction} \citep{BAIL}  \citep{Keep} \citep{MIR}. In particular, some algorithms filter dataset actions to extract the optimal behaviors and to ignore useless ones \citep{SoftRegularization} \citep{PolicyGuided} \citep{Discriminator} \citep{Replay}. Additionally, some algorithms employ policy regularization to implicitly avoid OOD actions \citep{BEAR} \citep{CR} \citep{AWR} \citep{AWAC} \citep{OneStep} \citep{Generalizing}. Another case is that some algorithms utilize uncertainty to constrain policy behavior \citep{Uncertatinty1} \citep{Uncertatinty2} \citep{Uncertatinty3}. These algorithms have achieved great success in offline settings. However, there are two shortcomings for policy regularization. On one hand they could never eliminate OOD actions completely. On the other hand they are not good enough for the offline dataset with random actions and low-performing trajectories.

The other tack is the value functions ($Q(s,a)$ and $V(s)$) regularization or constraints. These algorithms overestimate $Q$-value function with in-distribution actions and underestimate that with out-of-distribution actions \citep{CQL} \citep{IQL} \citep{MCQ} \citep{SQL} \citep{PenalizedQ} \citep{FisherDivergence} \citep{Algaedice}. They also have made significant breakthroughs in offline settings. However, two challenges block their improvement. Firstly, it is hard to distinguish between in-distribution actions and out-of-distribution ones without querying actions in datasets, even if a BC model is trained to produce in-distribution actions. Meanwhile, considering that $Q$-value function model is always a neural network, when algorithms overestimate some actions and underestimate the other actions by gradient descent, changed parameters may lead to unexpected effects due to the mutual influence of model parameters.

Recently, diffusion model is introduced into offline RL models to reinforce the policy expressive capability and they achieve great success in improving the performance \citep{diffusionQL} \citep{IDQL} \citep{SfBC} \citep{CEP} \citep{EDP} \citep{SRPO}.

Aspired from $Q$ regularization and diffusion policy, our algorithm implicitly underestimates all value functions to counteract overestimation phenomenon based on theoretical analysis, which can attenuate the mutual influence of model parameters. 
%Furthermore, expectile regression can assign $Q$-value function with in-distribution actions with a bit higher returns.

\section{Preliminaries}
%\subsection{Offline Reinforcement Learning}

The processing of Reinforcement Learning (RL) can be represented as Markov Decision Process (MDP) specified by a tuple: $\mathcal{M} = \left \langle \mathcal{S}, \mathcal{A}, \mathcal{P}, r, \rho _{0}, \gamma   \right \rangle $ where $\mathcal{S}$ is the state space, $\mathcal{A}$ is the action space, $\mathcal{P}$: $p({s}'|s,a)$ is the transition probability from the pair of state and action $(s,a)$ to the next state ${s}'$, $r(s,a) \in \mathcal{R}$ is the reward function, $\rho _{0}$ is the initial state distribution, and $\gamma$ is the discount factor. The goal of RL is to train a policy $\pi (\cdot |s,a)$ to maximize the expected accumulated long-term rewards as follows.
\begin{equation}
	\label{e15}
	J(\pi) = \mathbb{E} _{s_{0} \sim \rho_{0}(s), \,a_{t} \sim \pi(a|s_{t}), \, s_{t+1} \sim p(s|s_{t+1},a_{t+1})} \left [ \sum_{t=0}^{\infty } \gamma^{t}r(s_{t},a_{t})  \right ].    
\end{equation} 
The contraction bellman backup can be employed to estimate the long-term returns of $Q$-value function as the Eq. (\ref{e16}) for the corresponding policy $\pi$.
\begin{equation}
	\label{e16}
	\mathcal{T} Q^{\pi}(s,a) : =  r(s,a) + \mathbb{E}_{s'} \left [ \mathbb{E}_{a' \sim \pi(\cdot |s')} \left [ Q^{\pi}(s',a') \right ] \right ].       
\end{equation} 
If $\pi$ is the optimal policy, the corresponding optimal bellman operator is in the Eq. (\ref{e17}). And $Q$-value function is usually parameterized in the form of neural network.
\begin{equation}
	\label{e17}
	\mathcal{T} Q(s,a) : =  r(s,a) + \mathbb{E}_{s'} \left [ \max_{a' \in \mathcal{A}} \left [ Q(s',a') \right ] \right ].      
\end{equation} 
In the offline settings, it is infeasible to online interact with the environment to correct the $Q$-value function and policy model. The offline algorithms only learn the $Q$-value function and policy model by the previously collected dataset $\mathcal{D}$. Then mean square error (MSE) of temporal difference (TD) is minimized as the Eq. (\ref{e18}) to train $Q$-value function.
\begin{equation}
	\label{e18}
	\begin{split}
		&L_{mse}(\theta) \\
		&= \mathbb{E}_{(s,a,s')\sim \mathcal{D}} \left [ \left ( \mathcal{T}Q_{\theta}(s,a) -  Q_{\theta}(s,a) \right )^{2}  \right ]  \\
		&= \mathbb{E}_{(s,a,s')\sim \mathcal{D}} \left [ \left ( r(s,a) + \max_{a'\sim \mathcal{A}} Q_{{\theta}'}(s',a') -  Q_{\theta}(s,a) \right )^{2}  \right ],
	\end{split}     
\end{equation}
where $\theta$ is the parameter set of $Q$-value function and $Q_{{\theta}'}(s',a')$ is the target $Q$-value function without gradient backpropagation.

%Furthermore some of them can be estimated a bit higher because of expectile regression.
%Moreover, in the beginning of reward backpropagation the error gradually increases which harms the estimation of value functions. employ the expectile regression to realize it. In addition So $\iota$ in the quantile operator $q^{\iota}$ need to be less than $0.5$. takes $\iota$ quantile In the Equation (\ref{e12}) 
%We provide the new operator under deterministic dynamics for simplicity. As for stochastic dynamics, the same conclusion will be obtained. We show its proof under the deterministic dynamics. As for the stochastic dynamics, the similar proof procedure will provide the same property.
%behavior cloning (BC) baseline and 10\% BC that performs behavior cloning on the best 10.

\section{The Principle of the Proposed UDQL}

In this section, we introduce the design of offline reinforcement learning algorithms in detail. Firstly, we theoretically derive the overestimated quantity of value functions under Gumbel distribution. Then we propose the bellman operator and prove that it is a contraction operator to counteract the overestimated error. Furthermore, we employ a simple and easy for coding style to realize that operator. Finally, we build the deterministic diffusion policy and provide the general offline RL algorithm. %with\\without behavior cloning.

\subsection{Overestimated Error Analysis}

We analyze the estimation problem existing in the value function under deterministic dynamics for simplicity. If the environment is stochastic dynamics, the same conclusion will be obtained. If MSE loss is used, the Eq. (\ref{e31}) and (\ref{e32}) are viewed as location parameters of $\hat{V}(s)$ and $\hat{Q}(s,a)$ instead of the Eq. (\ref{e21}). We set $\hat Q(s,a)$ and $\hat V(s)$ for the $T$-th time-step to follow the Eq. (\ref{e5}).
\begin{equation}
	\begin{split}
		\label{e5}
		&\hat{V_T} (s_{T}) \sim \mathcal{G}\left ( \mathbb{L}^{\beta_{T}}_{a_{T} }\left [ \hat{Q}(s_{T},a_{T}) \right ] + \gamma_{e} \beta_{T} , \beta_{T} \right ) \\
		&\hat{Q_T} (s_{T} ,a_{T}) \sim \mathcal{G}\left ( r(s_{T} ,a_{T}) + \gamma \mathbb{L}^{\beta_{T+1}}_{a_{T+1}}\left [ \hat{Q}(s_{T+1},a_{T+1}) \right ] + \gamma_{e} \beta_{T}, \beta_{T}   \right ),
	\end{split}
\end{equation}
where $\beta_{T} = \gamma \beta_{T+1}$ and $\gamma$ is the discount factor.

Therefore, we can derive the quantity of overestimation error as \emph{Theorem 1}.

\emph{\textbf{Theorem 1}: when $\beta_{T} = \beta$ and $\hat{Q_t} (s_{t} ,a_{t})$ follow Eq. (\ref{e5}), the overestimated error of $\hat{Q_t} (s_{t} ,a_{t})$ in the $t$-th time-step of the trajectory is $(T-t+1) \gamma_{e} \gamma^{T-t} \beta$, where $\gamma$ is discount factor and $\gamma_{e}$ is Euler-Mascheroni constant}.

% \begin{equation}
	%   \begin{split}
		%   \label{e7}
		%   \tilde{Q} (s_{t} ,a_{t}) = r(s_{t} ,a_{t}) + \gamma \mathbb{L}^{\gamma^{T-t-1}\beta}_{a_{t+1}}\left [ Q(s_{t+1},a_{t+1}) \right ] + (T-t+1) \gamma_{e} \gamma^{T-t} \beta
		%   \end{split}
	% \end{equation}

% \emph{$Q(s_{t+1},a_{t+1})$ is the optimal $Q$-value function without overestimated error as the Equation (\ref{e21}) and $\tilde{Q}(s,a)$ is estimated $Q$-value function under MSE loss as the Equation (\ref{e3}) }  the Equation (\ref{e7}).
The proof of \emph{Theorem 1} are deferred to Appendix \ref{appenA1}. The theorem provides the bounding of the overestimation error of $Q$-value function.

Similarly, we can prove the overestimated error of $V(s_{t})$ under MSE loss as \emph{Theorem 2}.

\emph{\textbf{Theorem 2}: when $\beta_{T} = \beta$ and both $\hat{Q_t} (s_{t} ,a_{t})$ and $\hat{V_t} (s_{t})$ follow Eq. (\ref{e5}), the overestimated error of $V(s_{t})$ in the $t$-th time-step of the trajectory is $(T-t+2) \gamma_{e} \gamma^{T-t} \beta$, where $\gamma$ is discount factor and $\gamma_{e}$ is Euler-Mascheroni constant}.

The proof of \emph{Theorem 2} is provided in Appendix \ref{appenA2}. The theorem gives the bounding of the overestimation error of $V$-value function.

Based on \emph{Theorem 1} and \emph{Theorem 2}, we can build the relationship of $\tilde{Q}(s,a)$ and $\tilde{V}(s)$ and consistently they conform to the general value function formula without the overestimated error as \emph{Theorem 3} although both $\tilde{Q}(s,a)$ and $\tilde{V}(s)$ have overestimated errors.

\emph{\textbf{Theorem 3}: $\tilde{Q}(s_{t},a_{t})$ has the relationship with $\tilde{V}(s_{t+1})$ as Eq. (\ref{e10})}.
\begin{equation}
	\begin{split}
		\label{e10}
		\tilde{Q} (s_{t} ,a_{t}) = r(s_{t} ,a_{t}) + \gamma \tilde{V}(s_{t+1}).
	\end{split}
\end{equation}

%\begin{equation}
%	\nonumber
%	\tilde{Q} (s_{t} ,a_{t}) = r(s_{t} ,a_{t}) + \gamma \tilde{V}(s_{t+1}).
%\end{equation}
The proof of \emph{Theorem 3} is provided in Appendix \ref{appenA3}. It gives the common relationship between $Q$-value function and $V$-value function.

\textbf{Remark 2.0} \emph{Theorem 3} demonstrates that the overestimated error existing in the value function has inherent logic of self-consistency because the common value function relationship between $Q(s,a)$ and $V(s)$ still holds under its condition. As for the overestimated error function, we show its property as follows.

\emph{\textbf{Property 1}: As for the overestimated error function $f(x)=Cx\gamma^{x-b}$, where both $C=\gamma_{e}\beta$ and $b \in \{1,2\}$ are constants and $x = T-t \ge 0$,
	\begin{itemize}
		\item (1) $f(x)$ is the concave function;
		\item (2) $\lim\limits_{x \to \infty} f(x) = 0$.
	\end{itemize}
}

\emph{Property 1} indicate the feature of overestimated function and its proof is presented in Appendix \ref{appenB1}. From \emph{Property 1} we can get that the overestimated error quantity increases first and then decreases until that $T \to +\infty$ it fades to $0$. However, in the beginning the error gradually increases which remarkably damages the estimation of value functions. In particular, SARSA-TD approach is always used to estimate $Q(s,a)$ and $V(s)$ rather than Monte Carlo and it only considers the relationship of value functions of adjacent states, which hinder overestimated error to fade and promote it to grow.

\subsection{Underestimated Quantile Operator}

When MSE is used in estimating value functions, there is overestimation phenomenon existing in $Q(s,a)$ and $V(s)$. Although we have provided the specific error quantity in the above section, in practice it is intractable to estimate the exact quantity of overestimated error due to noises and insufficient samples. So we propose a practical and feasible solution to address the overestimation problem. That is, we present a novel Bellman operator that underestimates the value function and then we prove that the operator is a $\gamma$-contraction.

\emph{\textbf{Definition 1}: The underestimated quantile bellman operator is defined as Eq. (\ref{e12}),
	\begin{equation}
		\label{e12}
		\mathcal{T}^{\iota} Q^{\pi} = q^{\iota}\left ( r(s,a) + \gamma \mathbb{E}_{s'}\left [ \max_{a'\in \pi } Q^{\pi}(s',a') \right ] \right ),   
	\end{equation}  
	where $\pi$ is the current policy and $q^{\iota}(x)$ means taking the $\iota$ quantile of $(x)$ $0 \le \iota \le 1$.
}

The basic idea behind the proposed bellman operator is that underestimating the $Q$-value function is needed due to overestimation phenomenon. 

\textbf{Qualitative Analysis} The novel operator has two angles of meanings. On the one hand, essentially the $q^{\iota}$ operation is similar to reducing the quantity of discount factor $\gamma$ to $\iota \gamma$, which can accelerate overestimating error fading based on \emph{Propery 1}. That also weakens the effect of the current action on future rewards in $Q$-value function. On the other hand, the novel operator also underestimates the current returns in order to counteract the overestimated error.

\emph{\textbf{Property 2}: The quantile bellman operator $\mathcal{T}^{\iota}$ is $\iota \gamma$-contraction in the $\mathcal{L}_{\infty}$ norm.
}

The proof of \emph{Property 2} is provided in Appendix \ref{appenB2} and it demonstrates that the proposed bellman operator can make value function converge. 

In practice, we employ the expectile regression to realize the quantile operation, which inspired from IQL \citep{IQL}. We modify the loss function to estimate the current $Q$-value function to be less than the target as the Eq. (\ref{e14}).
\begin{equation}
	\label{e14}
	L(\theta ) = \mathbb{E}_{(s,a,{s}')\sim D,{a}'\sim \pi} \left [ L^{\tau}_{2} \left ( Q^{\theta } (s,a) - \left ( r(s,a) + Q^{{\theta}'}({s}',{a}') \right ) \right )   \right ],   
\end{equation} 
where $L^{\tau}_{2}\left ( u \right ) = \left | \tau - \mathrm {1} (u < 0) \right | u^{2}$; $\theta$ is the parameter set of $Q$-value function model; $\pi$ is the current policy.

Because we underestimate the current $Q(s,a)$, we set $\tau > 0.5$. In deep reinforcement learning, $Q$-value function is represented by a deep neural network with parameter set $\theta$. Then it is updated by $Q^{k+1} \gets \mathcal{T}^{\iota}Q^{k}$ where $k$ is the step number of gradient descent. The expectile regression is tightly related to the quantile regression \citep{IQL} \citep{Quantile}. Specifically, the target $Q$-value network has been built with parameters ${\theta}'$ and it is updated with lag and without gradient backpropagation. So in practice the Eq. (\ref{e14}) is employed to update $Q$-value network in the light of $Q^{k+1}_{\theta} \gets q^{\iota } \left ( r + Q^{k}_{{\theta}'} \right )$.

\subsection{Policy Learning and General Algorithm}

We utilize the conditional diffusion model as the parametric policy because it has more expressive representation capability. Its forward process is employed to training and its reverse process is used to action inference as the Eq. (\ref{e27}).
\begin{equation}
  \label{e27}
  \pi _{\phi}\left ( a | s \right ) = p_{\phi}\left ( a^{0:N}|s \right ) = p(a^{N})\prod_{n=1}^{N} p_{\phi}\left ( a^{n-1} | a^{n},s \right ),     
\end{equation} 
\noindent where $p(a^{N})$ is usually normal Guassian distribution $\mathcal{N} \left ( a^{N} | 0, I \right )$ and $p_{\phi}\left ( a^{n-1} | a^{n},s \right )$ is also modeled as Guassian distribution $\mathcal{N} \left ( a^{n-1} ; \mu_{\phi}(a^{n},s,n), \Sigma _{\phi}(a^{n},s,n) \right )$ \citep{DDPM}. The diffusion model loss is used as policy regularization in training process following \citep{diffusionQL} as shown in the Eq. (\ref{e28}).
\begin{equation}
  \label{e28}
  \mathcal{L}_{d}(\phi) = \mathbb{E}_{n\sim \mathcal{U},\ \epsilon \sim \mathcal{N}(0,\ I),\ (s,a) \sim \mathcal{D}  }\left (\left \|   \epsilon - \epsilon _{\phi}\left ( \sqrt{\bar{\alpha}_{n} }\ a + \sqrt{1 - \bar{\alpha}_{n} } \ \epsilon  , s, n \right )  \right \|^{2} \right )     
\end{equation} 
Furthermore, the diffusion policy learning objective is maximizing Q- function with policy regularization as the Eq. (\ref{e29}). Additionally, we use $\eta$ and $\zeta$ to weight deterministic policy loss and regularization because they can intuitively to express their responsibility although either of them with the learning rate can achieve the same effect not-intuitively. We detail the learning procedure of our UDQL in Algorithm \ref{algudql}.
\begin{equation}
  %\begin{split}
  \label{e29}
  \mathcal{L}\left ( \phi  \right ) = \eta \mathcal{L}_{q}(\phi) + \zeta \mathcal{L}_{d}(\phi) = - \eta * \mathbb{E}_{s\sim \mathcal{D} }\left ( Q\left ( s, \pi_{\phi}(s)  \right )  \right ) + \zeta * \mathcal{L}_{d}(\phi).
  %\end{split}
\end{equation} 

Our method simply underestimates $Q$-value a little by expectile regression, which can lead to improving the performance marginally in offline RL environment. Because common regression loss MSE will overestimate $Q$-value based on our theoretical analysis, especially in TD-style. The advantage of expectile regression that is used as the practical execution of underestimated operator is simplicity and ease for coding. 
%The disadvantage of expectile regression realizing underestimated operator is its training a little unstably.

\begin{algorithm}%[!htb] 
  \caption{UDQL Algorithm} 
  \label{algudql} 
  \begin{algorithmic}%[1] %这个1 表示每一行都显示数字
  %\REQUIRE ~~\\ %Input
  % \\
  %\\
  %\ENSURE ~~\\ %Output
  %$\hat{}$\\
  %\IF {not the end of current sentence}
  \STATE Initialize critic networks $Q_{\theta_1}$, $Q_{\theta_2}$ and actor network $\pi_{\phi}$ with random parameters
  \STATE Initialize target networks ${\theta}'_1 \gets \theta_1$, ${\theta}'_2 \gets \theta_2$, ${\phi}' \gets \phi$ and offline replay buffer $\mathcal{D}$
  \FOR {each iteration do}
  \STATE Sample a mini-batch $B= \{ ( s,a,r,s' ) \} \sim \mathcal{D}$
  \STATE Sample $\hat{a}' \sim \pi_{{\phi}'}(a'|s')$
  \STATE Update $Q_{\theta_1}$, $Q_{\theta_2}$ using $\mathcal{L}(\theta )$ from Eq. (\ref{e14})
  \STATE Sample $\hat{a} \sim \pi_{\phi}(a|s)$
  \STATE Update $\pi_{\phi}$ using $\mathcal{L}\left ( \phi \right )$ from Eq. (\ref{e29})
  \STATE Update ${\theta}'_{i}$, ${\phi}'$ by \\
   ${\theta }'_{i} = \rho {\theta }'_{i} + (1-\rho) \theta_{i}$, ${\phi }' = \rho {\phi }' + (1-\rho) \phi$
  
  %\STATE ENCODE 
  \ENDFOR
  
  %\RETURN $\widehat{F}$;
  \end{algorithmic}
\end{algorithm}

%\mathbb{E}_{s\sim \mathcal{D} }\left ( Q\left ( s, \pi_{\phi}(s)  \right )  \right ) + \zeta * \mathbb{E}_{n\sim \mathcal{U},\ \epsilon \sim \mathcal{N}(0,\ I),\ (s,a) \sim \mathcal{D}  }\left (\left \|   \epsilon - \epsilon _{\phi}\left ( \sqrt{\bar{\alpha}_{n} }\ a + \sqrt{1 - \bar{\alpha}_{n} } \ \epsilon  , s, n \right )  \right \|^{2} \right )
\section{Experiments}

In this section, we conduct extensive experiments to demonstrate the effectiveness and efficiency of our proposed algorithm on the D4RL benchmark \citep{d4rl}. Firstly, we provide the experimental setup and baselines. Then we compare our algorithm to existing state-of-the-art algorithms in the offline settings. 
%Finally, we present the qualitative analysis for the experimental results.

\subsection{Baselines and Experimental Setup}

\textbf{Baselines} Our algorithm adds expectile regression with underestimated $Q$-value function on the standard implementations of diffusionQL (DQL) \citep{diffusionQL}. Essentially, it is a variant of TD3 \citep{TD3} using the novel value estimation operation with diffusion policy. We compare our method to other state-of-the-art offline baselines, including BC\citep{BC}, BCQ \citep{BCQ}, CQL \citep{CQL}, One-step \citep{OneStep}, DT \citep{DT}, TD3+BC \citep{TD3BC}, IQL \citep{IQL}, AWAC \citep{AWR} \citep{AWAC}, diffusionQL \citep{diffusionQL}. Furthermore, we also compare our method to the state-of-the-art diffusion policy and $Q$ regularization offline RL models, including: SfBC \citep{SfBC}, IDQL \citep{IDQL}, QGPO \citep{CEP}, $\chi$-QL C and $\chi$-QL T \citep{XQL}, DiffusionQL \citep{diffusionQL}, SRPO \citep{SRPO}, SQL and EQL \citep{SQL} and most of them are recently proposed. All extensive baselines include various categories of offline approaches: value regularization, policy regularization or constraints, imitation learning, one-step, mixed methods and so on.
  
\textbf{Experimental Setup}
We evaluate our proposed algorithms on the D4RL benchmark, including: Gym-locomotion, Adroit, AntMaze and Kitchen tasks. The architecture of $Q$ networks and diffusion policy networks is kept unchanged in all experiment tasks that is the same with \citep{diffusionQL} \citep{EDP}, i.e., $3$ hidden layers MLP with Mish \citep{Mish} activation function and each layer with $256$ neural nodes. We use Adam \citep{Adam} as the optimizer to train $Q$ networks and policy networks. We exploit the GTX$3060$ GPU to implement the experiments. Following \citep{diffusionQL} \citep{EDP}, as for Gym-locomotion task the models are trained for $2000$ epochs, and with respect to Adroit, AntMaze and Kitchen tasks they are trained for 1000 epochs. Each epoch contains $1000$ iterations with the batch size of $256$. The evaluation normalized score is obtained at intervals of $50$ epochs \citep{diffusionQL} and it is not a frequent evaluation so we take the normalized score of the best-performing model (online model selection OMS) \citep{diffusionQL} as the evaluation metrics. Additionally, we find that the best-performing model can also perform very well in without-random-seed environment and furthermore as long as the best model has been obtained, it can solve the corresponding problem. The time-step of diffusion policy is set to $5$. The experimental scores are calculated by averaging $5$ random seeds and detailed logs are attached in supplementary materials.

\begin{table}[t]
	\caption{{\footnotesize Averaged normalized scores of UDQL against SOTA baselines on the D4RL benchmark. The UDQL scores are taken over 50 random rollouts (10 evaluations with 5 seeds) for Gym-locomotion tasks and over 500 random rollouts (100 evaluations with 5 seeds) for AntMaze, Adroit and Kitchen tasks. The results of the SOTA baselines are directly adopted from their papers. UDQL achieves the highest scores in 13 out of 16 tasks.}}
	\label{table1}
	\centering
	\tiny
	\setlength{\tabcolsep}{3pt}
	\begin{tabular}{l|llllllllll|l}
		\toprule
		% \multicolumn{2}{c}{Part} \\
		%\cmidrule(r){1-2}
		%\midrule
		Gym Tasks & BC & AWAC & Diffuser & MoREL & Onestep RL & TD3+BC & DT & CQL & IQL & DiffusionQL & UDQL \\
		\midrule
		halfcheetah-medium-v2 & 42.6 & 43.5 & 44.2 & 42.1 & 48.4 & 48.3 & 42.6 & 44.0 & 47.4 & 51.1 & \textbf{70.7} $\pm$ 1.0 \\
		hopper-medium-v2 & 52.9 & 57.0 & 58.5 & 95.4 & 59.6 & 59.3 & 67.6 & 58.5 & 66.3 & 90.5 & \textbf{100.9} $\pm$ 0.4 \\
		walker2d-medium-v2 & 75.3 & 72.4 & 79.7 & 77.8 & 81.8 & 83.7 & 74.0 & 72.5 & 78.3 & 87.0 & \textbf{88.6} $\pm$ 0.9 \\
		halfcheetah-medium-replay-v2 & 36.6 & 40.5 & 42.2 & 40.2 & 38.1 & 44.6 & 36.6 & 45.5 & 44.2 & 47.8 & \textbf{64.2} $\pm$ 1.6 \\
		hopper-medium-replay-v2 & 18.1 & 37.2 & 96.8 & 93.6 & 97.5 & 60.9 & 82.7 & 95.0 & 94.7 & 101.3 & \textbf{103.1} $\pm$ 0.3 \\
		walker2d-medium-replay-v2 & 26.0 & 27.0 & 61.2 & 49.8 & 49.5 & 81.8 & 66.6 & 77.2 & 73.9 & 95.5 & \textbf{98.7} $\pm$ 2.0 \\
		halfcheetah-medium-expert-v2 & 55.2 & 42.8 & 79.8 & 53.3 & 93.4 & 90.7 & 86.8 & 91.6 & 86.7 & 96.8 & \textbf{101.4} $\pm$ 1.2 \\
		hopper-medium-expert-v2 & 52.5 & 55.8 & 107.2 & 108.7 & 103.3 & 98.0 & 107.6 & 105.4 & 91.5 & \textbf{111.1} & \textbf{111.5} $\pm$ 0.7 \\
		walker2d-medium-expert-v2 & 107.5 & 74.5 & 108.4 & 95.6 & \textbf{113.0} & 110.1 & 108.1 & 108.8 & 109.6 & 110.1 & 112.1 $\pm$ 0.6 \\ 
		\midrule
		%\cmidrule(r){1-2}
		\midrule
		AntMaze Tasks & BC & AWAC & BCQ & BEAR & Onestep RL & TD3+BC & DT & CQL & IQL & DiffusionQL & UDQL \\
		\midrule
		antmaze-umaze-v0 & 54.6 & 56.7 & 78.9 & 73.0 & 64.3 & 78.6 & 59.2 & 74.0 & 87.5 & 93.4 & \textbf{97.6} $\pm$ 1.4 \\
		antmaze-umaze-diverse-v0 & 45.6 & 49.3 & 55.0 & 61.0 & 60.7 & 71.4 & 53.0 & 84.0 & 62.2 & 66.2 & \textbf{78.4} $\pm$ 7.6 \\
		\midrule
		
		%\cmidrule(r){1-2}
		\midrule
		Adroit Tasks & BC & SAC & BCQ & BEAR & BEAR-p & BEAR-v & REM & CQL & IQL & DiffusionQL & UDQL \\
		\midrule
		pen-human-v1 & 25.8 & 4.3 & 68.9 & -1.0 & 8.1 & 0.6 & 5.4 & 35.2 & 71.5 & \textbf{72.8} & 71.0 $\pm$ 8.2 \\ 
		pen-cloned-v1 & 38.3 & -0.8 & 44.0 & 26.5 & 1.6 & -2.5 & -1.0 & 27.2 & 37.3 & 57.3 & \textbf{62.9} $\pm$ 4.9 \\ 
		\midrule
		
		%\cmidrule(r){1-2}
		\midrule
		Kitchen Tasks & BC & SAC & BCQ & BEAR & BEAR-p & BEAR-v & AWR & CQL & IQL & DiffusionQL & UDQL \\
		\midrule
		kitchen-complete-v0 & 33.8 & 15.0 & 8.1 & 0.0 & 0.0 & 0.0 & 0.0 & 43.8  & 62.5 & 84.0 & \textbf{87.8} $\pm$ 6.7 \\
		kitchen-partial-v0 & 33.8 & 0.0 & 18.9 & 13.1 & 0.0 & 0.0 & 15.4 & 49.8 & 46.3 & \textbf{60.5} & 56.7 $\pm$ 2.3 \\
		kitchen-mixed-v0 & 47.5 & 2.5 & 8.1 & 47.2 & 0.0 & 0.0 & 10.6 & 51.0 & 51.0 & 62.6 & \textbf{67.5} $\pm$ 3 \\ 
		\bottomrule
	\end{tabular}
\end{table}

\subsection{Results on MuJoCo Datasets}

%Furthermore some of them can be estimated a bit higher because of expectile regression.  
%Moreover, in the beginning of reward backpropagation the error gradually increases which harms the estimation of value functions. employ the expectile regression to realize it. In addition So $\iota$ in the quantile operator $q^{\iota}$ need to be less than $0.5$. takes $\iota$ quantile In the Equation (\ref{e12}) 
%We provide the new operator under deterministic dynamics for simplicity. As for stochastic dynamics, the same conclusion will be obtained. We show its proof under the deterministic dynamics. As for the stochastic dynamics, the similar proof procedure will provide the same property.
%behavior cloning (BC) baseline and 10\% BC that performs behavior cloning on the best 10.

We compare our offline RL method with multiple state-of-the-art (SOTA) offline algorithms in D4RL benchmark in Table \ref{table1} and show averaged normalized scores in the form of mean $\pm$ deviation. The results of the prior algorithms are directly adopted from corresponding papers. We can see that our method outperforms almost all prior algorithms except $3$ tasks. Specifically, in the gym-locomotion task our algorithm nearly outperforms all state-of-the-art ones. Especially in the challenging halfcheetah tasks in which their scores are almost below $50$, our algorithm surpasses all of them with a large margin and nearly improves their scores by $40\%$. In the Adroit, AntMaze and Kitchen tasks, our method outperforms them marginally in most tasks. Even if in the tasks ours are surpassed, our scores are very close to theirs. Our model achieves the highest scores in 13 out of 16 tasks in Table \ref{table1} with little standard deviation.

Table \ref{table2} shows the performance of particular offline models based on diffusion policy and $Q$ regularization. Most of them are lastest proposed models and have improved performance. The Adroit task is excluded in Table \ref{table2} because it is lost in some other models. We can see that our method outperforms almost all prior algorithms except $3$ tasks. In the sub-optimal tasks of our model, it is other errors but not overestimation that dominate the returns and we don't provide extra solutions to addressing other errors. But our model also achieves the best scores in 11 out of 14 tasks in Table \ref{table2}.

\begin{table}[t]
	\caption{{\footnotesize Averaged normalized scores of UDQL against offline diffusion policy or $Q$ regularization models on the D4RL benchmark. The UDQL scores are taken over 50 random rollouts  (10 evaluations with 5 seeds) for Gym-locomotion tasks and over 500 random rollouts (100 evaluations with 5 seeds) for AntMaze and Kitchen tasks. The Adroit task is excluded because it is lost in some other models. The results of other models are directly adopted from their papers. UDQL achieves the highest scores in 11 out of 14 tasks.}}
	\label{table2}
	\centering
	\scriptsize
	\setlength{\tabcolsep}{3pt}
	\begin{tabular}{l|lllllllll|l}
		\toprule
		%\multicolumn{2}{c}{Part}                   \\
		\cmidrule(r){1-2}
		Dataset     & SfBC & IDQL & QGPO & $\chi$-QL C &$\chi$-QL T & DiffusionQL & SRPO & SQL& EQL & UDQL \\
		\midrule
		halfcheetah-medium-v2       & 45.9  & 49.7  & 54.1 & 47.7 & 48.3 &51.1 & 60.4 & 48.3 & 47.2 &\textbf{70.7 $\pm$ 1.0}\\
		hopper-medium-v2            & 57.1  & 63.1  & 98.0 & 71.1 & 74.2 & 90.5 & 95.5 & 75.5 & 74.6 &\textbf{100.9} $\pm$ 0.4 \\
		walker2d-medium-v2          & 77.9  & 80.2  & 86.0 & 81.5 & 84.2 & 87.0 & 84.4 & 84.2 & 83.2 &\textbf{88.6} $\pm$ 0.9 \\
		halfcheetah-medium-replay-v2& 37.1  & 45.1  & 47.6 & 44.8 & 45.2 & 47.8 & 51.4 & 44.8 & 44.5 &\textbf{64.2} $\pm$ 1.6 \\
		hopper-medium-replay-v2     & 86.2  & 82.4  & 96.9 & 97.3 & 100.7& 101.3& 101.2& 99.7 & 98.1 &\textbf{103.1} $\pm$ 0.3 \\
		walker2d-medium-replay-v2   & 65.1  & 79.8  & 84.4 & 75.9 & 82.2 &95.5 & 84.6 & 81.2 & 76.6 &\textbf{98.7} $\pm$ 2.0 \\
		halfcheetah-medium-expert-v2& 92.6  & 94.4  & 93.5 & 89.8 & 94.2 & 96.8 & 92.2 & 94.0 & 90.6 &\textbf{101.4} $\pm$ 1.2 \\
		hopper-medium-expert-v2     & 108.6 & 105.3 & 108.0& 107.1& 111.2& 111.1& 100.1& \textbf{111.8}& 105.5& \textbf{111.5} $\pm$ 0.7 \\
		walker2d-medium-expert-v2   & 109.8 & 111.6 & 110.7& 110.1& 112.7& 110.1& \textbf{114.0}& 110.0& 110.2& 112.1 $\pm$ 0.6 \\ 
		\midrule
		antmaze-umaze-v0         & 92.0 & 94.0 & 96.4 & 87.2  & 93.8 & 93.4 & 96.4 & \textbf{97.1} & 93.2 & \textbf{97.6} $\pm$ 1.4 \\
		antmaze-umaze-diverse-v0 & \textbf{85.3} & 80.2 & 74.4 & 69.17 & 82.0 & 66.2 & 74.4  & 82.1 & 65.4 & 78.4 $\pm$ 7.6 \\
		\midrule
		kitchen-complete-v0 & 77.9 & - & - & 72.5 & 82.4 & 84.0 & - & - & 70.3          & \textbf{87.8} $\pm$ 6.7 \\
		kitchen-partial-v0  & 47.9 & - & - & 73.8 & 73.7 & 60.5 & - & - & \textbf{74.5} & 56.7 $\pm$ 2.3 \\
		kitchen-mixed-v0    & 45.4 & - & - & 54.6 & 62.5 & 62.6 & - & - & 55.6          & \textbf{67.5} $\pm$ 3 \\ 
		\bottomrule
	\end{tabular}
\end{table}

We find that for double DQN we give two different $\tau$ to $Q1$ and $Q2$ networks that performs better, because diversity is very crucial for adapting to the environment. For the Gym-locomotion task we empirically set $\tau$ to $0.1$ and $0.2$; with respect to Adroit, AntMaze and Kitchen tasks, we empirically set it to $0.2$ and $0.3$. We find that different types of tasks need different appropriate $\tau$ due to different return mechanisms. Also we find that our model require larger quantities of gradient norm (gn) because expectile underestimation regression poses a hard challenge to Q networks. The hyperparameter setting is shown in the Appendix \ref{appenC1}.

\section{Conclusion and Future Work}
MSE is the often used loss tool in value function estimation of RL. However, in this paper, we find that there is a gap between MSE and the optimal value function, which will leads to overestimation phenomenon in value function estimation. We theoretically analyze the phenomenon and provide its upper bound. Then we propose a new bellman operator to counteract the overestimation. Also, we provide a simple style to execute the operator, expectile regression. Furthermore, we build a novel offline RL algorithm based on the novel operator and diffusion policy. Finally, experimental results demonstrate the effectiveness of our algorithm. In the future, we will explore a more efficient execution of underestimated operator to replace expectile regression. Also, we will explore a better model to address overestimation phenomenon in offline RL environment.

\section*{Limitations}

The limitation of our research is that there are also other factors that affect the performance heavily and we don't consider the mutual effect of all factors. Also, the proposed model is a bit sensitive to hyperparameters.

\clearpage

\medskip

{
\small
\bibliography{references}

}

%%%%%%%%%%%%%%%%%%%%%%%%%%%%%%%%%%%%%%%%%%%%%%%%%%%%%%%%%%%%

\newpage

\appendix
\begin{fontsize}{14pt}{20pt} % 第一个是字体大小，第二个是行距
	\textbf{Appendix}
\end{fontsize}

\section{Proof of Theoretical Results} 

\subsection{Proof of Theorem 1} \label{appenA1}

\textbf{Proof}: We can get the Eq. (\ref{e5}) based on the statistical characteristics of Gumbel distribution and the overestimated error is $\gamma_{e} \beta$. 
Then with regard to the error of $(t+1)$-th time-step $\gamma_{e} \beta_{t+1}$ we can obtain the overestimated error of $t$-th time-step as Eq. (\ref{e62}).
%%\begin{equation}
\begin{align}
	%\begin{split}  
	&\tilde{Q} (s_{t} ,a_{t}) = r(s_{t} ,a_{t}) + \gamma \mathbb{L}^{\beta_{t+1}}_{a_{t+1}}\left [ \tilde{Q}(s_{t+1},a_{t+1}) \right ] + \gamma_{e} \ast  \beta_{t} \nonumber \\
	&= r(s_{t} ,a_{t}) + \gamma \mathbb{L}^{\beta_{t+1}}_{a_{t+1}}\left [ r(s_{t+1},a_{t+1}) + \gamma \mathbb{L}^{\beta_{t+2}}_{a_{t+2}}\left [ \tilde{Q}(s_{t+2},a_{t+2}) \right ] + \gamma_{e} \beta_{t+1} \right ] + \gamma_{e} \beta_{t} \label{e61} \\
	&= r(s_{t} ,a_{t}) + \gamma \mathbb{L}^{\beta_{t+1}}_{a_{t+1}}\left [ r(s_{t+1},a_{t+1}) + \gamma \mathbb{L}^{\beta_{t+2}}_{a_{t+2}}\left [ \tilde{Q}(s_{t+2},a_{t+2}) \right ] \right ] + 2 \gamma_{e} \gamma^{T-t} \beta. \label{e62}
	%\end{split}
\end{align}
%\end{equation}

Given $\beta_{T} = \beta$ and the Eq. (\ref{e61}), we can get $\beta_{T-1} = \gamma \beta$ based on Extreme Value Theorem (EVT) and therefore the general relationship satisfies Eq. (\ref{e8}).
\begin{equation}
	\begin{split}
		\label{e8}
		\beta_{t} = \gamma \beta_{t+1} = \gamma^{T-t} \beta.
	\end{split}
\end{equation}
The Eq. (\ref{e62}) can be obtained when (\ref{e8}) is introduced into Eq.(\ref{e61}) .
Then we iteratively exploit Eq. (\ref{e62}) to compute the nested overestimated error and we can get Eq. (\ref{e7}).
\begin{equation}
	\begin{split}
		\label{e7}
		\tilde{Q} (s_{t} ,a_{t}) = r(s_{t} ,a_{t}) + \gamma \mathbb{L}^{\gamma^{T-t-1}\beta}_{a_{t+1}}\left [ Q(s_{t+1},a_{t+1}) \right ] + (T-t+1) \gamma_{e} \gamma^{T-t} \beta,
	\end{split}
\end{equation}
where $\gamma$ is discount factor; $\gamma_{e}$ is Euler-Mascheroni constant; $Q(s_{t+1},a_{t+1})$ is the optimal $Q$-value function without overestimated error as the Eq. (\ref{e21}); $\tilde{Q}(s,a)$ is estimated $Q$-value function under MSE loss as Eq. (\ref{e7}).
So by combining Eq. (\ref{e21}), the overestimated error is $(T-t+1) \gamma_{e} \gamma^{T-t} \beta$ in the $t$-th time-step. {\tiny$\blacksquare$}

\subsection{Proof of Theorem 2} \label{appenA2}

\textbf{Proof}:
According to the Eq. (\ref{e31}), we can obtain Eq. (\ref{e91}) and then we can employ the consequence of \emph{Theorem 1} to prove overestimated error as Eq. (\ref{e9}).
\begin{align}
	%\begin{split}
	%\label{e9}
	&\tilde{V} (s_{t}) = \mathbb{L}^{\beta_{t}}_{a_{t}} \left [ \tilde{Q}(s_{t},a_{t}) \right ] + \gamma_{e} \beta_{t} \label{e91} \\
	&= \mathbb{L}^{\beta_{t}}_{a_{t}} \left [ r(s_{t},a_{t}) + \mathbb{L}^{\beta_{t+1}}_{a_{t+1}}\left [ \gamma \tilde{Q}(s_{t+1},a_{t+1}) \right ]+ \gamma_{e}\beta_{t} \right ] + \gamma_{e} \beta_{t} \nonumber \\
	&= \mathbb{L}^{\gamma^{T-t}\beta}_{a_{t}}\left [ Q(s_{t},a_{t}) \right ] + (T-t+2) \gamma_{e} \gamma^{T-t} \beta \label{e9}. 
	%\end{split}
\end{align} 
\hfill{\tiny$\blacksquare$}

\subsection{Proof of Theorem 3} \label{appenA3}

\textbf{Proof}:
According to \emph{Theorem 1} we can get the following equation.
\begin{equation}
	\nonumber
	\tilde{Q} (s_{t} ,a_{t}) = r(s_{t} ,a_{t}) + \gamma \mathbb{L}^{\gamma^{T-t-1}\beta}_{a_{t+1}}\left [ Q(s_{t+1},a_{t+1}) \right ] + (T-t+1) \gamma_{e} \gamma^{T-t} \beta.
\end{equation}
According to \emph{Theorem 2} we can get the following equation.
\begin{equation}
	\nonumber
	\tilde{V}(s_{t+1}) = \mathbb{L}^{\gamma^{T-t-1}\beta}_{a_{t+1}}\left [ Q(s_{t+1},a_{t+1}) \right ] + (T-t+1) \gamma_{e} \gamma^{T-t-1} \beta.
\end{equation}
Therefore combining with the above equations we can obtain Eq. (\ref{e10}).  {\tiny$\blacksquare$}

\section{Proof of Property} 

\subsection{Proof of Property 1} \label{appenB1}

\textbf{Proof}:
The derivative of $f(x)$ is shown in Eq. (\ref{e11}) 
\begin{equation}
	\label{e11}
	{f}' (x) = C \left ( 1 +x \ln{\gamma}  \right ) \gamma^{x-b},
\end{equation}
where $ 0 < \gamma \le 1 $ is discount factor. 
It is trivial that ${f}'(x)$ monotonically decreases as $x$ increases. Therefore $f(x)$ is concave and when $x = -(\ln{\gamma})^{-1}$ $f(x)$ reaches its maximum value.

Because exponential function $\gamma^{x-b}$ has higher order increase/decrease than $x$, it is trivial that $\lim\limits_{x \to \infty} Cx\gamma^{x-b} = 0$ under the condition that $f(x)$ is concave. {\tiny$\blacksquare$}

\subsection{Proof of Property 2} \label{appenB2}

\textbf{Proof:}
For two arbitrary policy $\pi_{1}$ and $\pi_{2}$, we have
\begin{align}
	%\begin{split}
	%\label{e9}
	&\left \| \mathcal{T}^{\iota} Q^{\pi_{1}} - \mathcal{T}^{\iota} Q^{\pi_{2}} \right \| _{\infty}   \nonumber \\
	&= \max_{s,a} \left | q^{\iota}\left ( r(s,a) + \gamma \mathbb{E}_{s'}\left [ \max_{{a}'_{1} \in \pi_{1} } Q^{\pi_{1}}(s',{a}'_{1}) \right ] \right ) - q^{\iota}\left ( r(s,a) + \gamma \mathbb{E}_{s'}\left [ \max_{{a}'_{2}\in \pi_{2} } Q^{\pi_{2}}(s',{a}'_{2}) \right ] \right )  \right |  \nonumber \\
	&= \max_{s,a} \iota \gamma \left | \mathbb{E}_{s'}\left [ \max_{{a}'_{1} \in \pi_{1} } Q^{\pi_{1}}(s',{a}'_{1}) -  \max_{{a}'_{2} \in \pi_{2} } Q^{\pi_{2}}(s',{a}'_{2}) \right ]   \right | \nonumber \\
	&\le \iota \gamma \mathbb{E}_{s'} \left | \max_{{a}'_{1} \in \pi_{1} } Q^{\pi_{1}}(s',{a}'_{1}) - \max_{{a}'_{2} \in \pi_{2} } Q^{\pi_{2}}(s',{a}'_{2}) \right | \nonumber \\
	%&\le \iota \gamma \mathbb{E}_{s'} \max_{a'\in \mathcal{A}  } \left |  Q^{\pi_{1}}(s',a') -  Q^{\pi_{2}}(s',a') \right | \nonumber \\
	&\le \iota \gamma \left \|  Q^{\pi_{1}} -  Q^{\pi_{2}} \right \|_{\infty } \nonumber. 
	%\end{split}
\end{align}
\hfill{\tiny$\blacksquare$}

\section{Experimental Hyperparameters} \label{appenC1}

The experimental hyperparameter is shown in the Table \ref{table3} as following.

\begin{table}[h]
    \caption{{\footnotesize Experimental Hyperparameters}}
    \label{table3}
    \centering
    \scriptsize
    \setlength{\tabcolsep}{3pt}
    \begin{tabular}{l|lllllllll}
    \toprule
    %\multicolumn{2}{c}{Part}                   \\
    %\cmidrule(r){1-2}
    Dataset                      & $\tau$(Q1) & $\tau$(Q2) & lr & $\eta$ & $ \zeta $ & gn & num epochs & max q backup & batch size  \\
    \midrule
    halfcheetah-medium-v2       & 0.1 & 0.2 & 3e-4 & 1.0 & 0.005 & 100.0 & 2000 & False & 256 \\
    hopper-medium-v2            & 0.1 & 0.2 & 3e-4 & 1.0 & 0.5   & 100.0 & 2000 & False & 256 \\
    walker2d-medium-v2          & 0.1 & 0.2 & 3e-4 & 1.0 & 0.2   & 100.0 & 2000 & False & 256 \\
    halfcheetah-medium-replay-v2& 0.1 & 0.2 & 3e-4 & 1.0 & 0.005 & 2.0   & 2000 & False & 256 \\
    hopper-medium-replay-v2     & 0.1 & 0.2 & 3e-4 & 1.0 & 0.1   & 10.0  & 2000 & False & 256 \\
    walker2d-medium-replay-v2   & 0.1 & 0.2 & 3e-4 & 1.0 & 0.1   & 4.0   & 2000 & False & 256 \\
    halfcheetah-medium-expert-v2& 0.1 & 0.2 & 3e-4 & 1.0 & 1.0   & 7.0   & 2000 & False & 256 \\
    hopper-medium-expert-v2     & 0.1 & 0.2 & 3e-4 & 1.0 & 1.0   & 100.0 & 2000 & False & 256 \\
    walker2d-medium-expert-v2   & 0.1 & 0.2 & 3e-4 & 1.0 & 1.0    & 5.0  & 2000 & False & 256 \\ 
    \midrule
    antmaze-umaze-v0            & 0.2 & 0.3 & 3e-4 & 0.5 & 1.0    & 10.0 & 1000 & True & 256 \\
    antmaze-umaze-diverse-v0    & 0.2 & 0.3 & 3e-4 & 2.0 & 1.0    & 3.0  & 1000 & True & 256 \\
    \midrule
    pen-human-v1                & 0.2 & 0.3 & 6e-5 & 0.1 & 1.0    & 50.0 & 1000 & True & 256 \\
    pen-cloned-v1               & 0.2 & 0.3 & 3e-5 & 0.01 & 1.0   & 0.0  & 1000 & True & 256 \\
    \midrule
    kitchen-complete-v0         & 0.2 & 0.3 & 3e-4 & 0.005 & 1.0  & 9.0   & 1000 & False & 256 \\
    kitchen-partial-v0          & 0.2 & 0.3 & 3e-4 & 0.005 & 1.0  & 100.0 & 1000 & False & 256 \\
    kitchen-mixed-v0            & 0.2 & 0.3 & 3e-4 & 0.005 & 1.0  & 100.0 & 1000 & False & 256 \\ 
    \bottomrule
    \end{tabular}
\end{table}

%%%%%%%%%%%%%%%%%%%%%%%%%%%%%%%%%%%%%%%%%%%%%%%%%%%%%%%%%%%%

\end{document}